\documentclass[a4paper,11pt]{article}
\usepackage{sobrapo-template}
\usepackage[english]{babel}
\usepackage[utf8]{inputenc}
\usepackage{amsmath,amssymb}
\usepackage{graphicx}
\usepackage{mathtools}
\usepackage[ruled,vlined,linesnumbered]{algorithm2e}
\usepackage{float}
\usepackage{makecell}
\usepackage[titletoc,title]{appendix}
\usepackage[square,numbers]{natbib}
\usepackage{epstopdf}
\usepackage{cite}
\usepackage{hyperref}

\usepackage{indentfirst}

\SetKwFor{ForEach}{for each}{do}{end}

\addto\extrasenglish{%
}

\newcolumntype{x}[1]{>{\centering\let\newline\\\arraybackslash\hspace{0pt}}p{#1}}
\graphicspath{{img/}}

\title{An approach to dealing with missing values in heterogeneous data using \textit{k}-nearest neighbors}

\begin{document}
	
\maketitle

\author{
	\name{Davi E. N. Frossard}
	\institute{Department of Computer Science, Federal University of Espirito Santo.}
	\iaddress{Av. Fernando Ferrari, 514, Vitoria, CEP 29075-910, Espirito Santo, ES, Brazil}
	\email{davienf1@gmail.com}
}

\author{ 
	\name{Igor O. Nunes}
	\institute{Department of Computer Science, Federal University of Espirito Santo.} 
	\iaddress{Av. Fernando Ferrari, 514, Vitoria, CEP 29075-910, Espirito Santo, ES, Brazil}
	\email {igordeoliveiranunes@gmail.com}
}

\author{ 
	\name{Renato A. Krohling}
	\institute{Department of Production Engineering \& Graduate Program in Computer Science,} 
	\iaddress{Av. Fernando Ferrari, 514, Vitoria, CEP 29075-910, Espirito Santo, ES, Brazil}
	\email {krohling.renato@gmail.com}
}

\vspace{8mm}

%% =-=-=-=-=-=-=-=-=-=-=
%%       ABSTRACT
%% =-=-=-=-=-=-=-=-=-=-=
\begin{abstract}
	Techniques such as clusterization, neural networks and decision  making usually rely on algorithms that are not well suited to deal with missing values. However, real world data frequently contains such cases. The simplest solution is to either substitute them by a best guess value or completely disregard the missing values. Unfortunately, both approaches can lead to biased results. In this paper, we propose a technique for dealing with missing values in heterogeneous data using imputation based on the \textit{k}-nearest neighbors algorithm. It can handle real (which we refer to as crisp henceforward), interval and fuzzy data. The effectiveness of the algorithm is tested on several datasets and the numerical results are promising.
\end{abstract}

\newpage
%% =====================
%%     INTRODUCTION
%% =====================
\section{Introduction}
\label{introduction}

Missing data occurs in many research areas such as data mining, machine learning and statistics, which often use algorithms that are not well suited to deal with missing values. In decision making, for instance, the presence of missing values may lead to invalid evaluation of the criteria, therefore biasing the result towards an inappropriate decision, as shown by \citet{imputedecision}.

A data-set is considered incomplete if an attribute of a feature is not observed. Many factors can lead to unobserved values, such as machine fault on sampling data, human refusal to provide full information and censorship. \citet{coarsedata} have shown that data may also be neither perfectly present nor entirely missing, instead, it is coarsened by rounding, heaping, censoring or other factors.

According to \citet{rubin}, missing data can be categorized into three categories: (i) Missing completely at random (\textit{MCAR}), if the probability of missing value in a variable is independent of the variable itself and on any other variable on the set; (ii) Missing at random (\textit{MAR}), when the probability of a missing value on a sample \textbf{\textit{X}} is independent of \textbf{\textit{X}} but follows a pattern throughout the data-set, therefore it can be predicted based on other variables; (iii) Not missing at random (\textit{NMAR}), when the probability of missing values on \textbf{\textit{X}} depends solely on \textbf{\textit{X}} itself. Therefore, \textit{MCAR} and \textit{MAR} are recoverable whereas \textit{NMAR} is, that is, the former can be imputed based on observation of other samples in the data-set and maximization of the likelihood.

Handling missing data while keeping data reliability may not be possible by disregarding the missing values or substituting by a best guess value. Given that, many imputation algorithms based on both statistical analysis and machine learning have been proposed, as seen in \citet{imputationmethods}. They include multiple hot deck and mean regression imputation (\citet{regression}). Machine learning methods include self-organizing maps (\citet{selforganizing}), multilayer perceptrons (\citet{mlpimp}) and clusterization (\citet{knn}).

Introduced by \citet{nnrule}, the nearest neighbor (\textit{NN}) rule is a nonparametric classification method in which a sample point is assigned to the nearest set of previously classified points. In its general definition, the \textit{k}-nearest neighbors classifier (\textit{k}-NN), uses the \textit{k} most related patterns of a test set (\textit{TS}) to classify each of the patterns contained in the training set (\textit{TR}).

Although the \textit{k}-NN method's primary application belongs to supervised classification, it has also been widely used for the purpose of imputation (\citet{knn}; \citet{jonsson2004evaluation};  \citet{bras2007improving}). Particularly, the approach known as \textit{KNNimpute}, proposed by \citet{knn}, is broadly used for imputation in heterogeneous datasets, specifically with crisp data.

Motivated by its effectiveness, simplicity and wide applicability, this work is based on the usage of \textit{KNNimpute}. The goal of this paper consists in extending the standard \textit{k}-NN to handle heterogeneous data, i.e., crisp, interval and fuzzy. There exists a version of \textit{k}-NN for handling fuzzy datasets (\citet{knnfuzzy}). We focus specially in imputation in small datasets, where the task proves itself much harder than in bigger datasets given the higher propensity of over-fitting to the training data. As far as we know, there is no version of \textit{k}-NN for heterogeneous data.

In \autoref{Method} we provide a short review of the different data types with the respective distance measures used. In \autoref{extended}, an extended version of the \textit{k}-NN algorithm to handle heterogeneous data type is developed. Simulation results are presented in \autoref{results}. The paper ends up with conclusion and directions for future work in the area.

%% =====================
%%        METHOD
%% =====================
\section{Basic Definitions and the Standard \textit{k}-NN Algorithm}
\label{Method}

In this section we provide some basic definitions of the distance functions for different data types and review the standard \textit{k}-NN algorithm for data classification.

%% ---------------------
%%   DISTANCE MEASURES
%% ---------------------
\subsection{Distance Measures}
\label{distance measures}

In this section we present the basic definitions and distance functions of the data types, discussed with more details by \citet{distfunc}, that are used in this paper, which are crisp numbers, interval numbers and fuzzy sets.

%%       Crisp
%% ----------------
\subsubsection{Crisp Numbers}

A crisp number is composed of a single real value with no uncertainty attached to it. Given two scalars \textit{a} and \textit{b} then, the Euclidean distance is given by:

\begin{equation}
	\label{eq:dist_crisp}
	d(a, b) = \sqrt{(a - b)^2}
\end{equation}

%%     Interval
%% ----------------
\subsubsection{Interval Numbers}

An object $\bar{\textbf{a}}$ = $[a^L, a^U]$ where $\bar{\textbf{a}}$ is an interval number and the values $a^U$ and $a^L$ with $a^U \geq a^L$ are called the upper and lower bound, respectively. Given two interval numbers $\bar{\textbf{a}}$ and $\bar{\textbf{b}}$, the Euclidean distance between them is given by:

\begin{equation}
	\label{eq:dist_interval}
	d(\bar{\textbf{a}}, \bar{\textbf{b}}) = \frac{1}{2} \sqrt{(a^L - b^L)^2 + (a^U - b^U)^2}
\end{equation}

%%      Fuzzy
%% ----------------
\subsubsection{Fuzzy Sets}

A fuzzy set $\tilde{\textbf{a}}$ is described by a membership function $\mu_a: X \rightarrow [0,1]$. While there is no restrictions to the shape of the membership function, a common case is that of triangular membership functions, which are used throughout this paper.
Fuzzy sets with triangular membership functions are called triangular fuzzy numbers (TFN), denoted by $\tilde{\textbf{a}} = (a_1, a_2, a_3)$ with membership function given by:

\begin{equation}
	\label{eq:fuzzy_membership}
	\mu_a(x) = \begin{dcases*}
		\frac{x-a_1}{a_2 - a_1} & $a_1 < x < a_2$
		\\
		1 & $x =\ a_2$
		\\
		\frac{a_3 - x}{a_3 - a_2} & $a_2 < x < a_3$
		\\
		0 & otherwise
	\end{dcases*}
\end{equation}

Given two triangular fuzzy numbers  $\tilde{\textbf{a}}$ and  $\tilde{\textbf{b}}$, the distance between them is given by:
\begin{equation}
	\label{eq:dist_fuzzy}
	d(\tilde{\textbf{a}}, \tilde{\textbf{b}}) = \frac{1}{3} \displaystyle\sum_{i=1}^{3} \sqrt{(\textbf{a}_i - \textbf{b}_i)^2} 
\end{equation}

%% ---------------------
%%          KNN
%% ---------------------
\subsection{The Standard \textit{k}-Nearest Neighbor for Classification}
\label{knn}

In \textit{k}-Nearest Neighbors (\citet{knnclassic}), each instance is a point in a \textit{m}-dimensional Euclidean space. The algorithm is based on the observation that a sample can be classified in the same way as instances that share similar features, called the \textit{nearest neighbors}. Therefore, the nearest neighbors of an instance are those whose distance are the smallest. Let us assume that the distance is a function of each instance such as $d(X,Y)$. 
If we consider \textit{k} nearest neighbors, the assignment of a class to a sample is done by considering the class of each one of the \textit{k} nearest neighbors of the sample. The sample is then assigned to the class with most occurrences.
% Let us consider $T=t_1, t_2, ..., t_n$ the set of training samples and $S=s_1, s_2, ..., s_m$ the set of samples to be classified, then we have the procedure described in \autoref{KNNAlgorithm}:

\vspace{0.5cm}

%% =====================
%%   EXTENDED ALGORITHM
%% =====================
\section{The \textit{k}-NN Algorithm for Data Imputation}
\label{extended}

\subsection{The Standard \textit{k}-NNImpute}

Following similar principles to those discussed in \autoref{knn}, the \textit{k}-NNImpute algorithm (\citet{knn}), is an extension of the \textit{k}-NN algorithm used to impute missing values for crisp datasets. Using \autoref{eq:distancia}, one can calculate the distance between two instances ($X^i$ and $X^j$) containing $m$ values each. For imputation, such calculation must be executed for each sample with missing values and all samples which are candidates for imputation, i.e., instances which contain the value missing on the former. After finding the \textit{k} nearest neighbors by sorting the distances calculated with \autoref{eq:distancia}, the value can be imputed using \autoref{eq:imputar}, where the $w^j_v$ is the normalized weight of the instance calculated by \autoref{eq:peso}.

\begin{equation}
	\label{eq:distancia}
	d(X^i, X^j) = {\sqrt{ \frac{\sum\limits_{l=1}^{m} r^i_l . r^j_l . \sqrt{(x^i_l - x^j_l) ^2}}{\sum\limits_{l=1}^{m}r^i_l . r^j_l}}} \begin{dcases*}
		r^i_l = 0 & $x^i_l = NaN$
		\\
		r^j_l = 0 & $x^j_l = NaN$
		\\
		r^i_l = 1 & $x^i_l \neq NaN$
		\\
		r^j_l = 1 & $x^i_l \neq NaN$
	\end{dcases*}
\end{equation}

\begin{equation}
	\label{eq:imputar}
	x^j_l = \sum\limits_{v=i}^{k} w^j_v . x^v_l
\end{equation}

\begin{equation}
	\label{eq:peso}
	w^j_v = \frac{\frac{1}{d(x^j, x^v)}}{\sum\limits_{v=i}^{k} \frac{1}{d(x^j, x^v)}}
\end{equation}

Intuitively, \autoref{eq:distancia} changes the default L2 distance used in \textit{k}NN for classification to account for missing data, also making it so that instances that share the most number of common features have their distances reduced in comparison to samples that only have a few features in present in both. The \textit{k}-NNImpute algorithm is described in \autoref{KNNI}.
\vspace{0.2cm}

\begin{algorithm}
	\vspace{5px}
	\KwIn{Incomplete data-set $X_{D}$ \newline
		Number of nearest neighbors $k$}
	
	\KwOut{Imputed data-set $X'_{D}$}
	
	\vspace{10px}
	
	\ForEach{instance $X^j \in X_{D}$ with missing values}{
		\ForEach{missing value $X^i_l (l=1,2,...,m)$}{
			Find the \textit{k} nearest neighbors using \autoref{eq:distancia} with known values in position \textit{l}
			
			\ForEach{instance $x^v (v=1,2,...,k)$ closest to $x^j$}{
				Calculate the weight $w^j_v$ using \autoref{eq:peso}
			}
			
			Estimate $x^j_l$ value using \autoref{eq:imputar}
		}	
	}
	Return $X'_D$ 
	
	\caption{\textit{k}-NNImpute algorithm (\citet{knn})}
	\label{KNNI}
\end{algorithm}
\vspace{0.5cm}
In \textit{KNNimpute}, the set of closest neighbors is calculated for each instance with missing values. In this way, the set of neighbors can also contain missing values, as long as they are not on the same attribute. This could make the distance smaller for instances with more missing values. However, \autoref{eq:distancia} prevents that by not only considering the values itself, but how many of them the instances have in common.

\subsection{The Extended \textit{k}-NNImpute for Heterogeneous Data}

In order to be able to impute missing values in heterogeneous datasets, the distance function in \autoref{eq:distancia} must adapt to the data being processed. We assume the data-set is a matrix consisting of multiple data types, but they remain the same throughout each column. 
The heterogeneous method consists in calling the right distance function (among the ones discussed in \autoref{distance measures})  according to the column of origin. In this way, \autoref{eq:distancia} turns to:

\begin{equation}
	\label{eq:distanciaheterogeneous}
	\resizebox{.9\hsize}{!}{$d(X^i, X^j) = {\sqrt{ \frac{\sum\limits_{l=1}^{m} r^i_l . r^j_l . d(x^i_l, r^j_l)}{\sum\limits_{l=1}^{m}r^i_l . r^j_l}}} \begin{dcases*}
		r^i_l = 0 & $x^i_l = NaN$
		\\
		r^j_l = 0 & $x^j_l = NaN$
		\\
		r^i_l = 1 & $x^i_l \neq NaN$
		\\
		r^j_l = 1 & $x^i_l \neq NaN$
		\end{dcases*} \begin{dcases*}
		d(x^i_l, r^j_l) = \autoref{eq:dist_crisp} & class($x^i_l,r^j_l$) = Crisp
		\\
		d(x^i_l, r^j_l) = \autoref{eq:dist_interval} & class($x^i_l,r^j_l$) = Interval
		\\
		d(x^i_l, r^j_l) = \autoref{eq:dist_fuzzy} & class($x^i_l,r^j_l$) = Fuzzy
		\end{dcases*}$}
\end{equation}

Following this approach, it is easy to support more data types, as long as there is a known distance function for them. However, one must be careful not to use a distance function which results in distances with magnitudes superior to that of the other function, as that would bias the results for that specific data type. The distance functions in this paper have been thoroughly tested with respect to this property.

%% =====================
%%        RESULTS
%% =====================
\section{Simulation Results}
\label{results}

In order to evaluate the error in the method, values are removed at random from a data-set in a way that no line has more than one missing value or all the missing values fall in the same column. Values are then imputed and the mean square error between the original data-set ($\textbf{X}$) and the imputed set ($\textbf{X'}$) is calculated using \autoref{eq:MSE}.

\begin{equation}
	\label{eq:MSE}
	MSE(X',X) = \frac{\sqrt{X^2-X^{'2}}}{length(X)}
\end{equation}

%% ---------------------
%% HOMOGENEOUS data-setS
%% ---------------------
\subsection{Homogeneous datasets}
\label{homogeneous}

First we evaluate the algorithm using datasets consisting of a single data type. In this paper we evaluate Crisp, Interval and Fuzzy datasets.

%%       Crisp
%% ----------------
\subsubsection{Crisp data-set}
\begin{figure}[H]
	\begin{center}
		\includegraphics[width=\columnwidth]{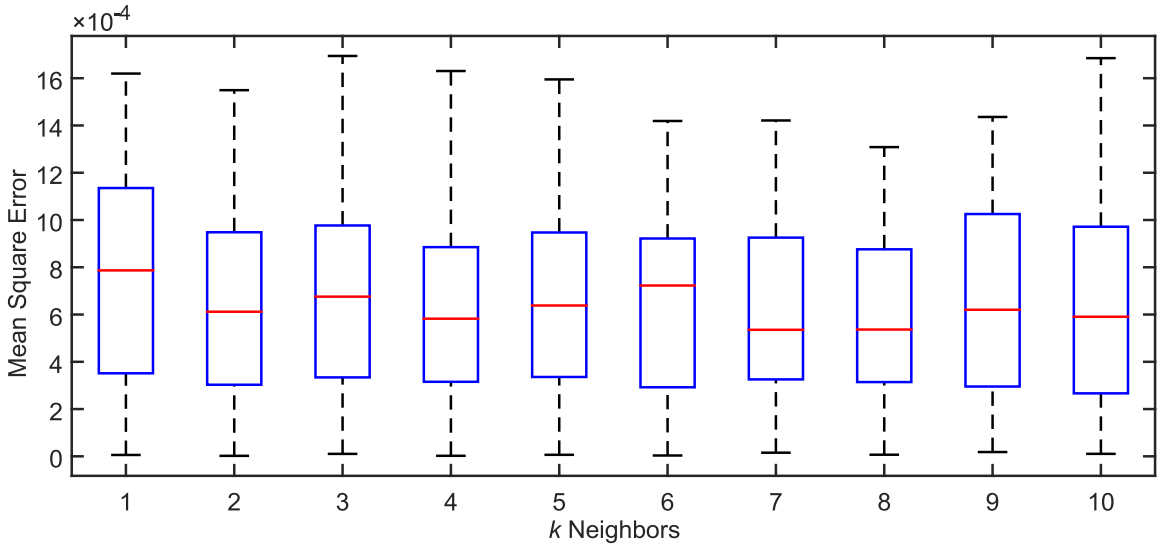}
		\caption{Imputation results for a crisp data-set}
		\label{fig:bmk1}
	\end{center}
\end{figure}

Imputing values missing at random in a data-set consisting of weather observations grouped on a 80x4 crisp matrix (available in \citet{jcmb}), where the number of NaNs ranged from 1 to 80 (25\% missing) one obtains the box-plot in terms of mean square error shown in \autoref{fig:bmk1}.

We notice that the best results occur for \textit{k}=7, and \textit{k}=8  with a mean square error (MSE) of $5.7$x$10^{-4}$ and the worst result for \textit{k}=1 with a MSE of $7.9$x$10^{-4}$, which indicates a small deviation in the results showing that there is evidence that the algorithm is robust with regards to the chosen number of neighbours (\textit{k}) considered during imputation.

%%     Interval
%% ----------------
\subsubsection{Interval data-set}
\begin{figure}[H]
	\begin{center}
		\includegraphics[width=\columnwidth]{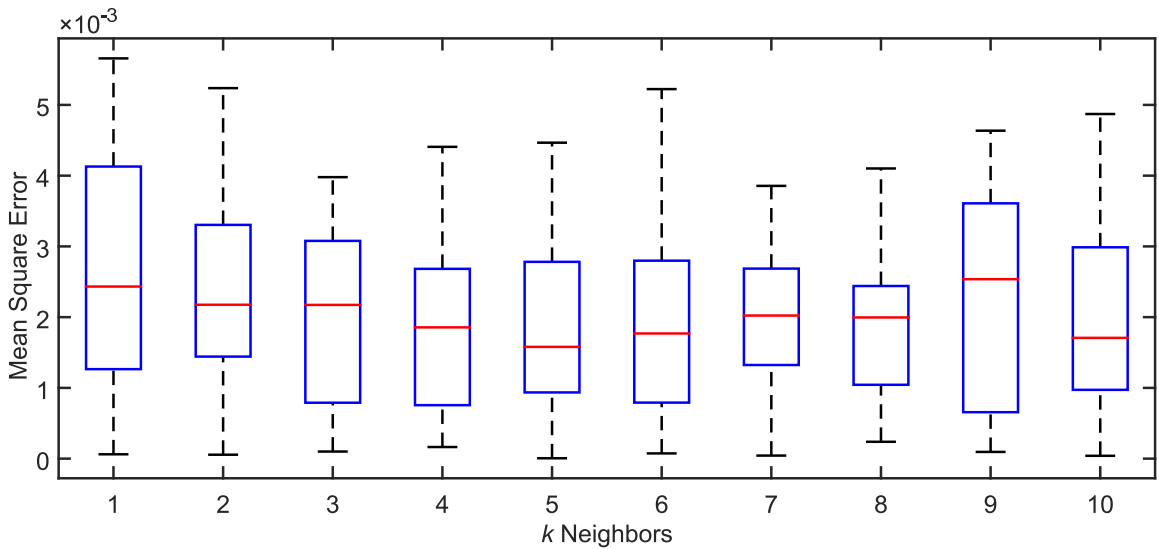}
		\caption{Imputation results for an interval data-set}
		\label{fig:bmk2}
	\end{center}
\end{figure}

Next, we use interval data, consisting of measures of bats (made available by \citet{bats}) and the number of NaN ranging from 1 to 21, one obtains the box-plot in terms of mean square error shown in \autoref{fig:bmk2}.

We can observe that the best results occur for \textit{k}=5 with a value of MSE of $1.6$x$10^{-3}$ and the worst result for \textit{k}=9 with a value of MSE of $2.5$x$10^{-3}$. The results reveals  that the algorithm is robust with regard to the number of  neighbors considered.

%%     Fuzzy
%% ----------------
\subsubsection{Fuzzy data-set}
\begin{figure}[H]
	\begin{center}
		\includegraphics[width=\columnwidth]{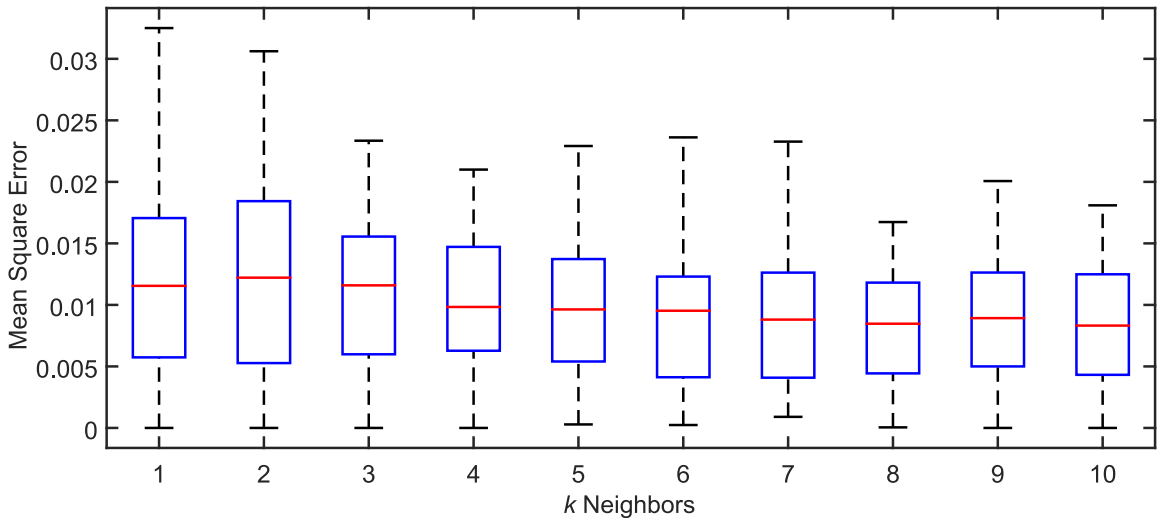}
		\caption{Imputation results for a fuzzy data-set}
		\label{fig:bmk3}
	\end{center}
\end{figure}

Using fuzzy data, consisting of characteristics of Taiwanese teas (made available by \citet{taitea}) and the number of NaNs ranging from 1 to 70, one obtains the results shown in the box-plot in \autoref{fig:bmk3}.

The box-plot shown in \autoref{fig:bmk3} shows that the best results in terms of MSE occurs at \textit{k}=10 with a value of $7.5$x$10^{-3}$ and the worst at \textit{k}=2 with a value of $1.2$x$10^{-2}$. The algorithm kept the trend that points to robustness with regard to the amount of neighbors considered.

%% ---------------------
%%    heterogeneous data-setS
%% ---------------------
\subsection{Heterogeneous Datasets}
\label{heterogeneous}

In this subsection we evaluate the algorithm with regard to heterogeneous datasets, especially small ones, where imputation becomes much harder.

%%     Case 1
%% ----------------
\subsubsection{Case Study 1}
To better illustrate the algorithm, we run an iteration considering two nearest neighbors (\textit{k}=2) for the data-set shown in \autoref{tab:htab2}, a decision making matrix made available by \citet{htab2}. Let us refer to it as M.

\begin{table}[H]
	\centering
	\begin{tabular}{|c|c|c|}
		\hline
		0.5891 & {[}0.31623, 0.94868{]} & {[}0.455842, 0.569803, 0.683763{]} \\\hline
		0.5624 & {[}0.55470, 0.83205{]} & {[}0.371391, 0.557086, 0.742781{]} \\\hline
		0.5802 & {[}0.55470, 0.83205{]} & {[}0.491539, 0.573462, 0.655386{]} \\\hline
	\end{tabular}
	\caption{Normalized decision matrix}
	\label{tab:htab2}
\end{table}

Suppose the fuzzy data at $M_{3,3}$ is missing, as presented in \autoref{tab:htab2_nan}.

\begin{table}[H]
	\centering
	\begin{tabular}{|c|c|c|}
		\hline
		0.5891 & {[}0.31623, 0.94868{]} & {[}0.455842, 0.569803, 0.683763{]} \\\hline
		0.5624 & {[}0.55470, 0.83205{]} & {[}0.371391, 0.557086, 0.742781{]} \\\hline
		0.5802 & {[}0.55470, 0.83205{]} & NaN \\\hline
	\end{tabular}
	\caption{Decision matrix with missing value}
	\label{tab:htab2_nan}
\end{table}

Computing the distances using \autoref{eq:distanciaheterogeneous} one obtains a distance of 0.2661 to line 1 and 0.0945 to line 2. We can now compute the normalized weight using \autoref{eq:peso}, which yields 0.2620 (line 1) and 0.7380 (line 2). Finally, we can impute the value according to \autoref{eq:imputar} as $0.2620M_{1,3} + 0.7380M_{2,3}$, resulting in $M_{3,3} = [0.3935, 0.5604, 0.7273]$, whose distance to the original value is of 0.0718 and a matrix mean square error of 0.0080.

\begin{figure}[H]
	\begin{center}
		\includegraphics[width=\columnwidth]{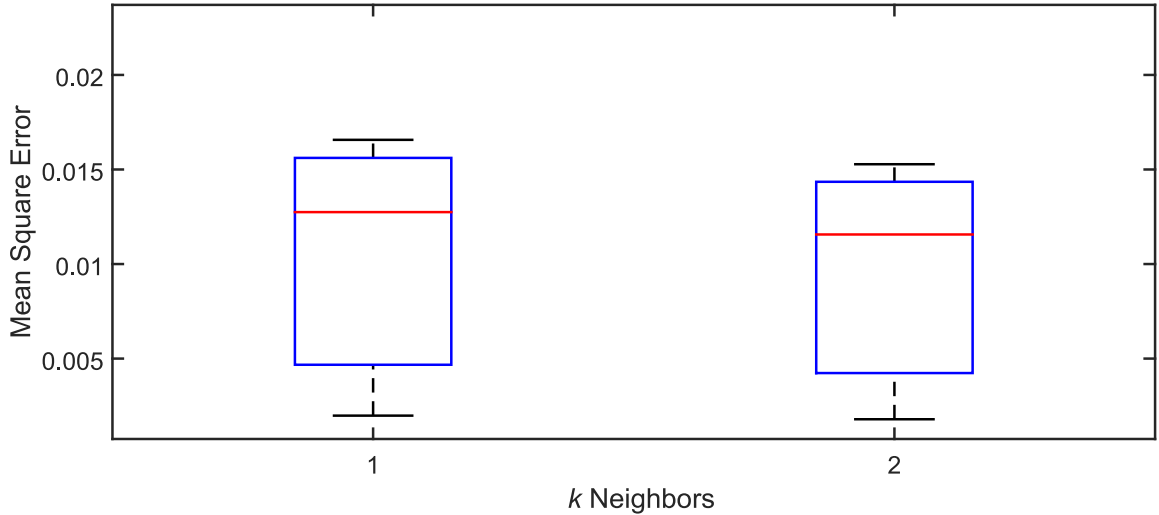}
		\caption{Imputation results for the matrix in \autoref{tab:htab2}}
		\label{fig:bmk4}
	\end{center}
\end{figure}

Running the benchmark with the number of NaNs ranging from 1 to 3, one obtains the box-plot in terms of MSE shown in \autoref{fig:bmk4}.

We can observe in \autoref{fig:bmk4} promising results for imputation of heterogeneous data, with both instances having a MSE of about $1.2$x$10^{-2}$. These results are especially interesting given the size of the matrix and how fast it deteriorates with the increase of NaN. The results obtained by using the algorithm keeps the MSE in the same range as those shown in the case of homogeneous datasets.

%%     Case 2
%% ----------------
\subsubsection{Case Study 2}
\label{hc2}

\begin{figure}[H]
	\begin{center}
		\includegraphics[width=\columnwidth]{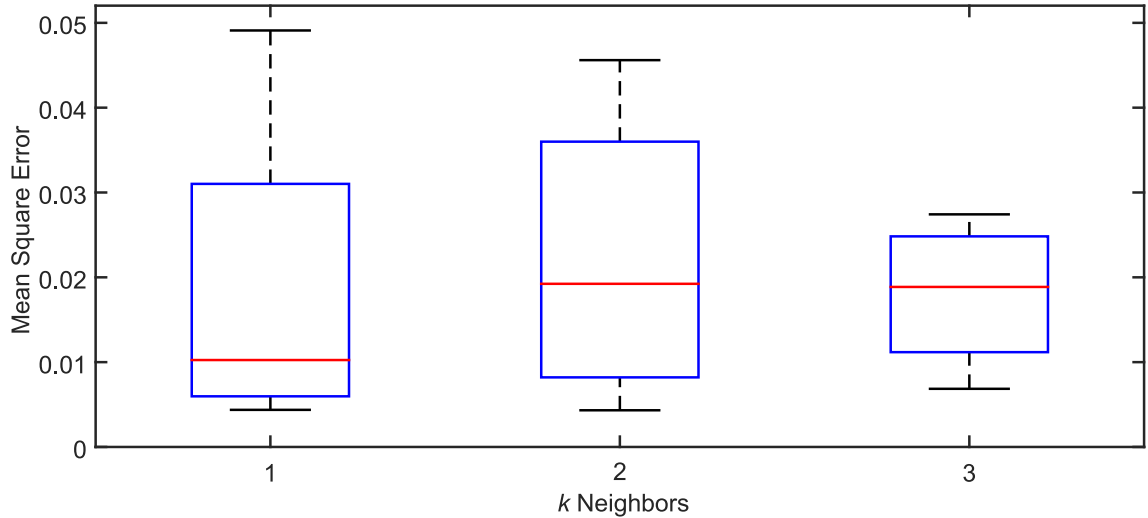}
		\caption{Imputation results for the matrix in \autoref{mctab1}}
		\label{fig:bmk5}
	\end{center}
\end{figure}

Applying the algorithm to a second benchmark involving a MCDM problem as presented in \autoref{mctab1} by \citet{htab1}, one obtains the results in terms of MSE as shown in \autoref{fig:bmk5}.

\begin{table}[H]
	\centering
	\begin{tabular}{|c|c|c|c|}
		\hline
		0.47 & [0.32, 0.48, 0.71] & [0.52, 0.67, 0.87] & [0.40, 0.55] \\\hline
		0.58 & [0.16, 0.29, 0.47] & [0.26, 0.37, 0.52] & [0.41, 0.58] \\\hline
		0.42 & [0.49, 0.67, 0.94] & [0.39, 0.52, 0.70] & [0.37, 0.54] \\\hline
		0.51 & [0.32, 0.48, 0.71] & [0.26, 0.37, 0.52] & [0.50, 0.69] \\\hline
	\end{tabular}
	\caption{Multi-criteria decision making matrix}
	\label{mctab1}
\end{table}

As expected, we notice in \autoref{fig:bmk5} that the algorithm presents a better behavior in terms of MSE with a more sizable data-set, providing  the best value of $1$x$10^{-2}$ for \textit{k}=1, and the worst value of $1.9$x$10^{-2}$ for \textit{k}=2. 

%%     Case 3
%% ----------------
\subsubsection{Case Study 3}

\begin{figure}[H]
	\begin{center}
		\includegraphics[width=\columnwidth]{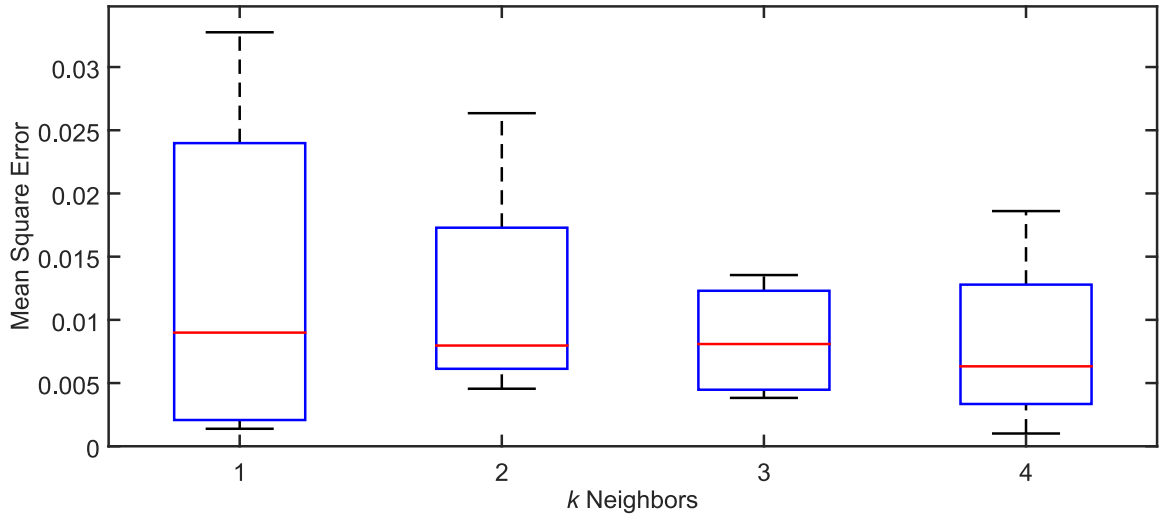}
		\caption{Imputation results for the matrix in \autoref{mctab2}}
		\label{fig:bmk6}
	\end{center}
\end{figure}

We consider now the MCDM matrix shown in \autoref{mctab2} (\citet{htab2}). The application of the algorithm provides the box-plot  shown in \autoref{fig:bmk6}.

\begin{table}[H]
	\centering
	\begin{tabular}{|c|c|c|c|}
		\hline
		0.45 & [0.60, 0.80] & [0.42, 0.57, 0.71] \\\hline
		0.41 & [0.37, 0.93] & [0.27, 0.53, 0.80] \\\hline
		0.48 & [0.32, 0.95] & [0.46, 0.57, 0.68] \\\hline
		0.43 & [0.55, 0.83] & [0.37, 0.56, 0.74] \\\hline
		0.46 & [0.20, 0.98] & [0.49, 0.57, 0.66] \\\hline
	\end{tabular}
	\caption{Multi-attribute decision matrix}
	\label{mctab2}
\end{table}

Using a data-set of about the same size as the one used in the previous case, but with more lines rather than columns (allowing more NaN and neighbours due to the benchmark mechanism, previously explained in \autoref{results}), the algorithm provided similar results, as depicted in \autoref{fig:bmk6}. The best result in terms of MSE obtained was $5.1$x$10^{-3}$ for \textit{k}=4 and the worst value of $5.8$x$10^{-3}$ for \textit{k}=1.

%% =====================
%%      CONCLUSIONS
%% =====================
\section{Conclusions}
\label{conclusions}

Since the \textit{k}-NN is a well established algorithm for classification, it has been extended to data imputation. Based on distance functions for the different data types used, the \textit{k}-NN impute has been modified and good results in terms of MSE were obtained. First, the algorithm was tested for homogeneous data types, i.e., crisp, interval and fuzzy data. Next, in order to test the approach it was necessary to use datasets from clusterization and decision making since there are no widely known benchmarks for imputation with heterogeneous datasets. Imputation results for heterogeneous data in terms of MSE shows insensitivity with respect to the numbers of neighbors used.

For future studies we suggest the development of a way to find out the optimal number of neighbors in order to produce the best imputation results and also the expansion to other data types. 	

\section*{Acknowledgments}
\label{acknowledgments}

D. E. N. Frossard would like to thank the brazilian agency CNPq for the scholarship under grant nr. 103863/2015-0. R. A. Krohling would also like to thank the financial support of the brazilian agency CNPq under grant nr. 303577/2012-6. 

\newpage
\bibliography{artigo}
\bibliographystyle{sbpo}

\end{document}